\pdfoutput=1

\documentclass[11pt]{article}

\usepackage[final]{acl}

\usepackage{times}
\usepackage{latexsym}

\usepackage[T1]{fontenc}

\usepackage[utf8]{inputenc}

\usepackage{microtype}

\usepackage{inconsolata}

\usepackage{graphicx}
\usepackage{tabularx}
\usepackage{booktabs}
\usepackage{adjustbox}
\usepackage{fontawesome}
\usepackage{amstext}
\usepackage{amsmath}
\usepackage{amssymb}
\usepackage{microtype}
\usepackage{multirow}
\usepackage{algorithm}
\usepackage{multicol}
\usepackage{algpseudocode}
\usepackage{amsfonts}
\usepackage{makecell}
\usepackage{colortbl} 
\usepackage{marvosym}
\usepackage{pifont}
\usepackage{hyperref}
\definecolor{mygreen}{RGB}{0,160,0}

\definecolor{myred}{RGB}{178,34,34}

\title{Meta-Cognitive Analysis: Evaluating Declarative and Procedural Knowledge in Datasets and Large Language Models}

\author{Zhuoqun Li${}^{1,3}$, Hongyu Lin${}^{1}$, Yaojie Lu${}^{1}$, Hao Xiang${}^{1,3}$, Xianpei Han${}^{1, 2}$, Le Sun${}^{1,2,}$\thanks{Corresponding Author}\\
${}^{1}$Chinese Information Processing Laboratory ~ ${}^{2}$State Key Laboratory of Computer Science \\
Institute of Software, Chinese Academy of Sciences, Beijing, China\\
${}^{3}$University of Chinese Academy of Sciences, Beijing, China \\
{\tt \{lizhuoqun2021,hongyu,luyaojie,xianghao2022,xianpei,sunle\}@iscas.ac.cn}}

\begin{document}
\maketitle

\begin{abstract}
Declarative knowledge and procedural knowledge are two key parts in meta-cognitive theory, and these two hold significant importance in pre-training and inference of LLMs. However, a comprehensive analysis comparing these two types of knowledge is lacking, primarily due to challenges in definition, probing and quantitative assessment. In this paper, we explore from a new perspective by providing ground-truth knowledge for LLMs and evaluating the effective score. Through extensive experiments with widely-used datasets and models, we get conclusions: (1) In most tasks, benefits from declarative knowledge are greater than those from procedural knowledge. (2) Profits of procedural knowledge are larger than declarative knowledge only in reasoning tasks with simple logic. (3) As pre-training progresses and size increases, model ability to utilize both  kinds of knowledge significantly improves, but in different speed. We do detailed analysis for the findings and this can provide primary guidance for evaluation and enhancement of large language models.
\end{abstract}

\section{Introduction}

Recent advancements in large language models (LLMs) have been noteworthy, with models such as GPT4 \cite{openai2023gpt4}, Llama \cite{touvron2023llama} and Vicuna \cite{vicuna2023} leading the way. These models are capable of solving various NLP tasks through an autoregressive approach and have show impressive performance \cite{Ye2023ACC,Bang2023AMM}. This evolution of LLMs has push NLP into a new era, moving away from traditional task-specific pre-train fine-tuning paradigm \cite{Zhao2023ASO}.

According to metacognitive theories~\cite{brown1987metacognition,jacobs1987children}, there are  two kinds of knowledge that critically contributes to the cognition of human beings: \emph{declarative knowledge} and \emph{procedural knowledge}. Declarative knowledge refers to ``knowing that''. It encompasses facts, concepts, and information that can be explicitly verbalized or described  \cite{ryle1945knowing,ryle2009concept}. For instance, knowing the capital of France is Paris or understanding the concept of gravity is declarative knowledge. Procedural knowledge, on the other hand, refers to ``knowing how''. It pertains to skills and procedures that we might not be able to verbalize explicitly but can demonstrate through action \cite{ryle1945knowing,ryle2009concept}. For example, knowing how to ride a bike, play a musical instrument, or swim are all forms of procedural knowledge. This type of knowledge is typically acquired through practice and repetition and is believed to involve the basal ganglia and cerebellum in the brain. Declarative and procedural knowledge are at the core of human cognition, underpinning our ability to function, learn, and adapt in various environments.

For large language models, many previous work has confirmed the existence of declarative and procedural knowledge in them. For declarative knowledge, it has been observed that LLMs are encoded with declarative knowledge across many domains~\cite{zhong2023agieval,huang2023ceval}, which is learned from the pre-training on vast portions of Internet. For procedural knowledge,  many researches have found that LLMs exhibit an understanding of processes or sequences of actions, to an extent. For instance, they can generate code snippets, describe step-by-step instructions, or help with problem-solving in certain domains, which significantly contribute to their ability to resolve complicated tasks~\cite{Gao2022PALPL,wei2023chainofthought,zhou2023leasttomost}.

Despite acknowledgment of their existence and significance within LLMs \cite{wei2023chainofthought,shi2023replug,fu2023specializing}, there is a marked absence of systematic exploration into how  these two kinds of knowledge exert on the capabilities of LLMs. This oversight considerably hampers our comprehensive understanding of large model dynamics and effective improvement of LLM abilities in certain tasks, domain and scenarios. A potential explanation for this absence stems from inherent challenges associated with formulating and probing these two kinds of knowledge and quantitatively assessing their implications on LLMs, respectively because of the black box nature of LLMs. Therefore, it is challenging to determine whether declarative or procedural knowledge is more critical for enhancement of current LLMs, as well as the impact of different types of knowledge on various models, tasks, and training phases. The lack of such studies significantly hampers efforts to target improvements in model capabilities.

\begin{figure}[t]
    \centering
    \includegraphics[width=\linewidth]{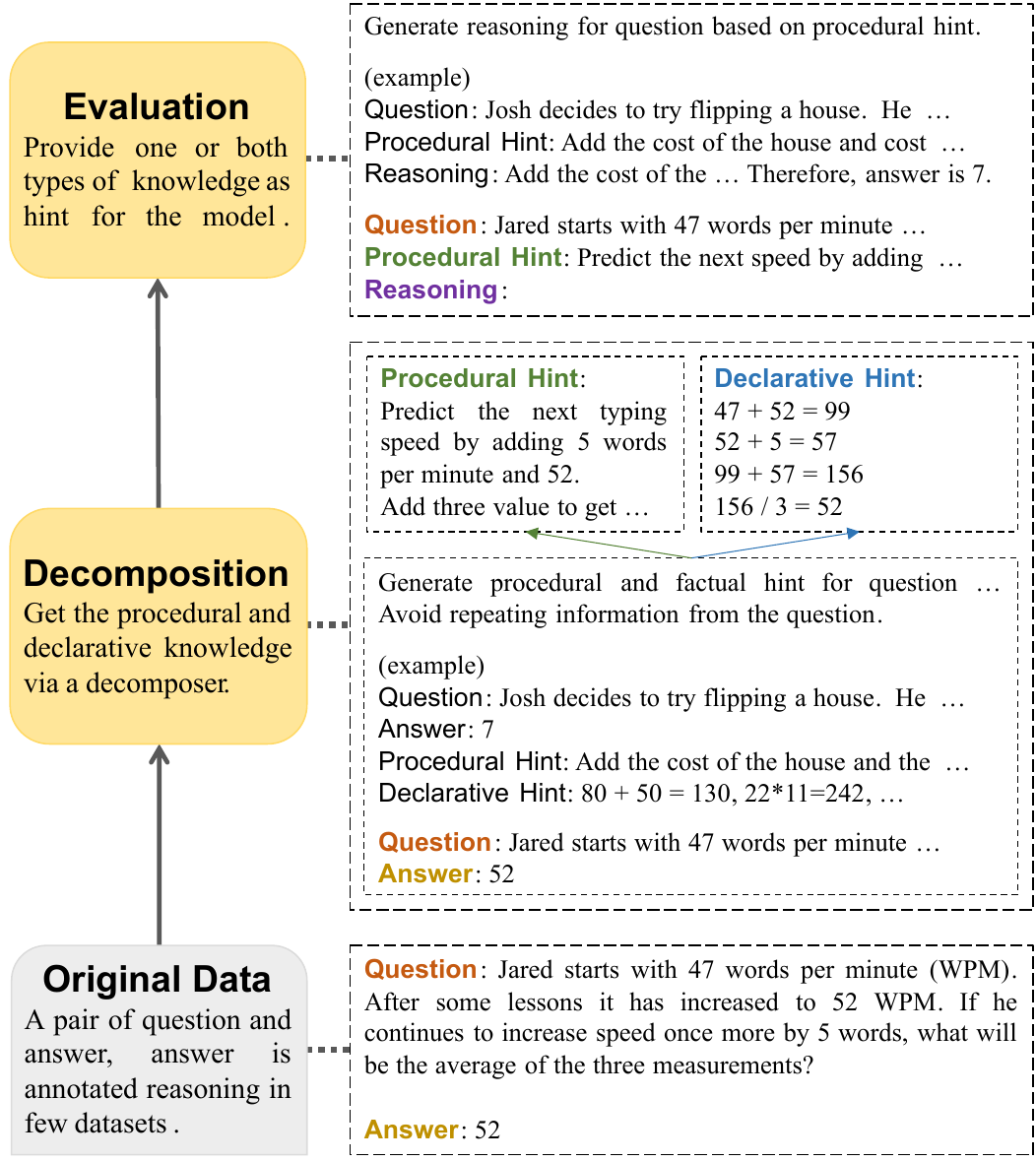} 
    \caption{The illustration of overall processing. First decompose original reasoning of the question to procedural knowledge and declarative knowledge, then evaluate models by providing one or both type of knowledge after the question.}
    \label{fig:processing}
\end{figure}

In this paper, we investigate the impact of declarative and procedural knowledge on datasets and large language models from a fresh perspective. Instead of directly probing these types of knowledge from LLMs, we examine how introducing ground truth of these two knowledge types affects the model performance on specific tasks. For this purpose, we design an exploration method based on in-context knowledge injection. By providing the necessary ground truth declarative or procedural knowledge in the question input to the LLM, we compare the model performance with and without this knowledge. This helps us understand the potential benefits of adding specific types of knowledge to LLMs. In this setup, the performance results can be roughly seen as the maximum capability of LLMs for a given task when they are not missing certain knowledge.

Specifically, our method primarily uses an in-context approach, providing the LLMs with three types of information specific to the question: 1) declarative hints: This type of information includes all the declarative knowledge needed to solve the current problem; 2) procedural hints: This type provides step-by-step plans of how to solve the problem; 3) combined hints: This information includes both of the above hints. It can be seen as the model maximum ability to use and combine the provided information to complete the task when given all the necessary details. We conduct experiments on 32 openly available large language models and 13 evaluation datasets that cover kinds of different tasks including math, commonsense and reasoning\footnote{Our detailed source codes are openly available at \url{https://github.com/Li-Z-Q/meta-cognitive-analysis}}. From our experiments, we find that:

\begin{itemize}
\setlength{\itemsep}{0.8pt}
    \item In most tasks, benefits from declarative knowledge are greater than those from procedural knowledge.
    \item Profits of procedural knowledge are larger than  declarative knowledge only in reasoning tasks with simple logic.
    \item As pre-training progresses and size increases, model ability to utilize both  kinds of knowledge significantly improves, but in different speed.
\end{itemize}
\section{Related Work}

Many works do illustrate declarative and procedural knowledge is important in LLM training and inference. LLMs can use chain of thought \cite{wei2023chainofthought,wang2023selfconsistency,zhou2023leasttomost} to solve complex tasks. Retrieval augmented LLMs \cite{ram2023incontext, shi2023replug} can use knowledge from knowledge bases or Internet to bolster model accuracy. LLMs can also be used to construct knowledge graph \cite{Cohen2023CrawlingTI}. In addition, some works try to inject declarative knowledge \cite{Kang2023KnowledgeAugmentedRD} or procedural knowledge \cite{fu2023specializing} during model training. However, there is no comprehensive analysis for these two types of knowledge in datasets and LLMs.

\section{Meta-Cognitive Analysis Method}

\begin{table*}[t]
  \centering
    \begin{tabular}{llcccc}
    \toprule
    \textbf{Dataset} & \textbf{Description} & \textbf{Procedural} & \textbf{Declarative} & \textbf{Combined} & \textbf{Delta} \\
    \midrule
    GSM8K &    Elementary Arithmetic     & \textbf{5.14} & 2.15 & 6.68 & \textbf{-2.99} \\
    MultiArith &    Elementary Arithmetic & \textbf{5.15} & 3.70  & 7.69 & \textbf{-1.45} \\
    CommonsenseQA      & Commonsense  & 1.17 & -0.06 & 1.45 & \textbf{-1.23} \\
    ARC-Easy    &  Commonsense    & 0.41 & 2.05 & 2.87 & 1.64 \\
    ARC-Challenge   &  Commonsense   & 0.47 & 3.00   & 3.27 & 2.53 \\
    TruthfulQA     & Commonsense  & -0.01 & 2.61 & 2.50  & 2.62 \\
    MMLU-STEM  &    Understanding   & 0.72 & 0.80  & 1.26 & 0.08 \\
    MMLU-Humanities    &   Understanding  & 0.59 & 1.53 & 2.11 & 0.94 \\
    MMLU-Social   &   Understanding   & 0.64 & 3.46 & 3.56 & 2.82 \\
    MMLU-Other    &   Understanding  & 0.06 & 3.05 & 3.49 & 2.99 \\
    MATH-1  & High School Mathematics  & 0.89 & 3.36 & 4.62 & 2.47 \\
    MATH-2  &  High School Mathematics     & 1.58 & 4.62 & 6.19 & 3.04 \\
    MATH-3   & High School Mathematics    & 1.33 & 4.52 & 6.07 & 3.19 \\
    \bottomrule
    \end{tabular}%
  \caption{Procedural, declarative and combined score of all datasets. The delta is how much larger the declarative score is than procedural score, it shows that procedural score is larger than declarative score only in GSM8K, MultiArith and CommonsenseQA. In addition, procedural scores of GSM8K and MultiArith are much larger than other datasets.}   
  \label{table:dataset}%
\end{table*}%




Procedural knowledge and declarative knowledge are two important aspects shared by most tasks. Exploring the difficulty of testing tasks and the capabilities of models from these two perspectives is of great significance for improving the training and testing of models. However, these two aspects, procedural knowledge and declarative knowledge, are coupled together in testing tasks. It requires a good method to decouple and quantify these two aspects, 
in order to analyze capabilities of models and the difficulty of test data.

In this paper, we address above challenges by a knowledge decomposition and injection method. Firstly, we provide clear definitions of procedural and declarative knowledge, and utilize GPT4\footnote{\url{https://platform.openai.com/docs/models/gpt-4-and-gpt-4-turbo}} to decouple the original reasoning into these two types of knowledge. Then, during the evaluation process, we provide questions and knowledge related to one aspect as hints to the model, observing the performance improvement of the model under these prompts. Subsequently, we transform these improvement value to scores and then do quantitative analysis on models and tasks.

\subsection{Decomposition of Declarative and Procedural Knowledge}


In order to quantify and analyze procedural and declarative knowledge, we first define these two types and then use GPT4 to decouple the original reasoning process into procedural knowledge and declarative knowledge.

Specially, we define declarative knowledge as fundamental facts crucial for resolving questions, each fact is independent with one another. Conversely, procedural knowledge embodies a generalized strategy essential for solving tasks, without any specific declarative details. Figure~\ref{fig:processing} shows one example, explaining the distinctions between declarative and procedural knowledge. 

With clear define of procedural and declarative knowledge, leveraging remarkable capabilities of GPT4, we do decomposition and get declarative and procedural knowledge for different tasks. In detail, by providing two examples for GPT4 (each example including a question, answer, and corresponding declarative knowledge and procedural knowledge), we let GPT4 decouple the two aspects of knowledge for unannotated question-answer data.

\subsection{Evaluation via In-context Hint Knowledge Injection}



After we obtain the decoupled data with procedural knowledge and declarative knowledge, we proceed to provide the model with a specific type of knowledge as a hint during the evaluation process. By observing the performance improvement resulting from these hints, we can further analyze the model capabilities in procedural and declarative aspects, the difficulty of tasks, and how effectively the hints are utilized.

Specially, during evaluation, models are given question with declarative knowledge, procedural knowledge, a combination of both, or neither, in contextual format to generate reasoning. One model input example  of procedural type is shown in Figure~\ref{fig:processing}, other types are similar with it. In a certain dataset-model pair, define $e_{o}$ is the original error rate, $e_{p}$ is the error rate after giving procedural knowledge, the procedural score is:
$$score_{p}=\frac{e_{o}-e_{p}}{e_{o}}$$
This score means effect of procedural knowledge. Via this metric, we can also get $score_{d}$ for declarative knowledge and $score_{c}$ for combination of both knowledge. With all dataset-model pairs score, to get a model score, we do average across all datasets and vice versa. 

\subsection{Datasets and Models}


To derive more general conclusions regarding procedural and declarative aspects, our experimental dataset encompasses varying degrees of knowledge and reasoning difficulty, our model incorporates diverse pre-training processes and sizes. 

In terms of datasets, for mathematics, we select easy mathematical datasets such as GSM8K \cite{cobbe2021training} and MultiArith \cite{roy2016solving}, and hard mathematical dataset MATH \cite{MATH}—of which we opt for levels 1, 2, and 3, excluding levels 4 and 5 due to their extreme complexity. For commonsense reasoning, we choose datasets including CommonsenseQA \cite{talmor2019commonsenseqa}, ARC-Easy \cite{clark2018think}, ARC-Challenge \cite{clark2018think}, and TruthfulQA \cite{lin2022truthfulqa}. Additionally, we choose MMLU \cite{hendrycks2021measuring} benchmark, which assesses many LLMs across four subcategories: humanities, social, STEM, and other. Note that while we focus on these selected datasets, our method is general and can be applied across all kinds of datasets.

\begin{figure}[t]
\centering
\includegraphics[width=\linewidth]{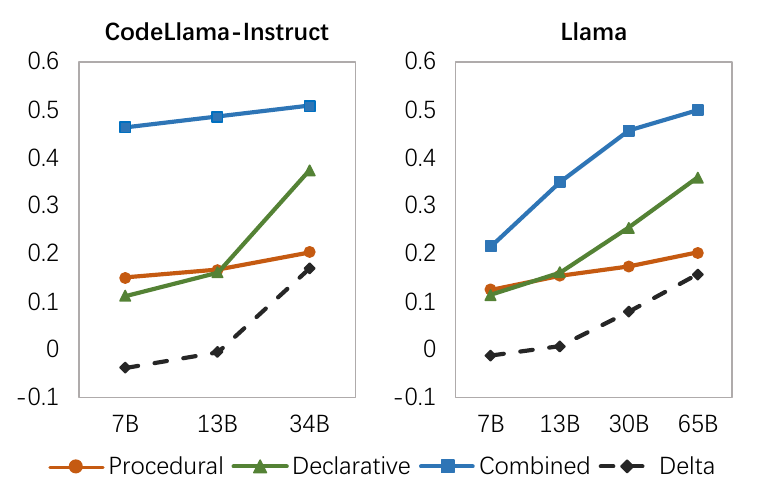} 
\caption{Procedural, declarative and combined score of different size models. The black line is the difference between procedural and declarative score. Above figure shows that the ability to utilize both knowledge becomes stronger as the model size increases, with different improvement rate.}
\label{fig:size}
\end{figure}

In terms of large language models, we select widely-used models including Llama \cite{touvron2023llama}, Vicuna-v1.3 \cite{vicuna2023}, Llama-2 \cite{touvron2023llama2}, Llama-2-Chat \cite{touvron2023llama2}, Vicuna-v1.5 \cite{zheng2023judging}, Vicuna-v1.5-16K \cite{zheng2023judging}, CodeLlama-Instruct \cite{rozire2023code}, and GPT3.5 API \footnote{\url{https://platform.openai.com/docs/guides/text-generation/chat-completions-api}}. Furthermore, we assess 11 checkpoints from Baichuan-2-7B \cite{Yang2023Baichuan2O}, each separated by an interval of 220 steps. Note that we only show a portion of models result in Figure~\ref{fig:size} and Figure~\ref{fig:baichuan}.


It is noteworthy that in some datasets, declarative hints may inherently encompass procedural elements due to sequencing. To address this, we collect noise declarative hints, mixing them with declarative hints, followed by randomization. 
\section{Findings}

\paragraph{Finding 1.} \textbf{\emph{In most tasks, benefits from declarative knowledge are greater than those from procedural knowledge.}}

As shown in Table~\ref{table:dataset}, we calculate the procedural score, declarative score, combined score and difference between procedural and declarative score, we observe that declarative score is larger than procedural score in 9 datasets. Thus we get conclusion that benefits from declarative knowledge is greater in most tasks.

\begin{figure}[t]
\centering
\includegraphics[width=\linewidth]{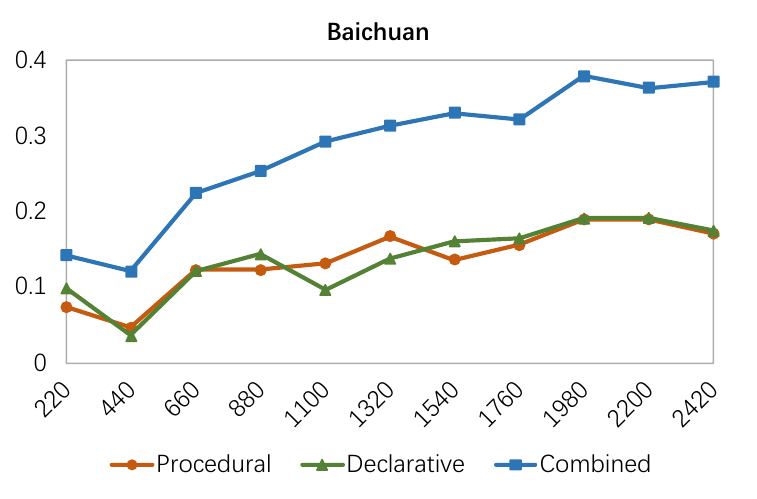} 
\caption{Scores of different checkpoints. The 220 means Baichuan-2-7B-00220, the model after 220 steps pre-training. It shows that the ability to utilize both type of knowledge becomes stronger as the pre-training step increases.}
\label{fig:baichuan}
\end{figure}
\paragraph{Finding 2.} \textbf{\emph{Procedural knowledge profits are larger than  declarative knowledge only in reasoning tasks with simple logic.}}

As shown in Table~\ref{table:dataset}, it shows that in simple mathematical datasets such as GSM8K and MultiArith, and basic commonsense reasoning dataset like CommonsenseQA, benefits from procedural knowledge are more than other datasets. This meets our expectations, as tasks like mathematics and commonsense reasoning often require more logical knowledge, commonly covered within procedural knowledge, hence golden procedural knowledge can improve a lot in these tasks.

On the other hand, for tasks with more complex logic, such as high school mathematics (MATH), we do not observe a significantly higher benefit from procedural knowledge than declarative knowledge, and gains from procedural knowledge are lower compared with simpler logic tasks. We speculate that this may be due to the limited capabilities of our test models. When the logic of question becomes overly complex, the model might struggle to understand and utilize logical information from procedural knowledge, resulting in the introduction of procedural knowledge not providing larger additional benefits.

\paragraph{Finding 3.} \textbf{\emph{As pre-training step and size increases, model ability to utilize both  kinds of knowledge significantly improves, but in different speed.}}

To find out effects of different model size and pre-training step, we do experiments in Llama, Code-Llama-Instruct and Baichuan model. Specifically, to observe impacts of model size in knowledge utilization, we compare performance of Llama 7B, 13B, 30B, 65B, and Code-Llama-Instruct 7B, 13B, 34B. To observe effects of pre-training step, we examine the  performance of Baichuan-2-7B at pre-training step from 220 to 2420.

In terms of model size, models with more parameters show clear improvement in capturing both declarative and procedural knowledge, as shown in Figure~\ref{fig:size}. Simultaneously, improvement in capturing declarative knowledge is significantly higher than procedural knowledge. This indicates that larger models are easier to utilize  external declarative information, while relevant procedural abilities might more rely on model inherent capabilities.

In terms of pre-training steps, as shown in Figure~\ref{fig:baichuan}, it shows a steady improvement in both type of knowledge as pre-training progresses. Note that basic performance of the model is continuously improving with more pre-training steps, we find that even above this baseline, the benefits of introducing additional knowledge continue to increase. This suggests that benefits of deeper training likely go beyond just enhancing the model’s internal knowledge capabilities, mainly improving the model ability to utilize knowledge.
\section{Conclusion}

In this paper, we conduct a comprehensive analysis of declarative and procedural knowledge, via a novel perspective. For each pair of dataset and model, we provide ground truth knowledge and then evaluate this knowledge effect score. Our experiments get insightful conclusions regarding the significance of these two types of knowledge across diverse datasets, and  efficacy with which different models utilize them. These findings provide primary guidance in enhancing evaluation and improvement processes of LLMs.

\section*{Acknowledgements}
We sincerely thank all anonymous reviewers for their insightful comments and valuable suggestions. This work is supported by the Strategic Priority Research Program of Chinese Academy of Sciences under Grant XDA27020200 and the Natural Science Foundation of China (No. 62122077 and 62106251).

\bibliography{custom}

 \end{document}